\pdfoutput=1

\documentclass[11pt]{article}

\usepackage[preprint]{acl}
\usepackage{graphicx}

\usepackage{booktabs}

\usepackage{times}
\usepackage{latexsym}
\usepackage{amsmath}
\usepackage{amssymb}
\usepackage{amsthm}
\usepackage[english]{babel}
\usepackage{makecell}
\usepackage{algorithm}
\usepackage{algorithmic}

\newcommand{\A}{A}

\newcommand{\LL}{\mathcal{L}}

\newcommand*{\E}{\mathbb{E}}
\usepackage[T1]{fontenc}

\usepackage[utf8]{inputenc}
\usepackage{microtype}

\usepackage{inconsolata}

\usepackage{graphicx}

\title{LoRA$^2$ : Multi-Scale Low-Rank Approximations for Fine-Tuning Large Language Models}

\author{
 \textbf{Jia-Chen Zhang\textsuperscript{1}},
 \textbf{Yu-Jie Xiong\textsuperscript{1}\thanks{Corresponding author.}},
 \textbf{He-Xi Qiu\textsuperscript{1}},
 \textbf{Dong-Hai Zhu\textsuperscript{1}},
 \textbf{Chun-Ming Xia\textsuperscript{1}}
\\
 \textsuperscript{1}School of Electronic and Electrical Engineering, Shanghai University of Engineering Science, \\
 333 Longteng Road, Songjiang District, Shanghai, China
\\
 \small{
   \textbf{Correspondence:} \href{mailto:xiong@sues.edu.cn}{xiong@sues.edu.cn}
 }
}

\begin{document}
\maketitle
\begin{abstract}
Fine-tuning large language models (LLMs) with high parameter efficiency for downstream tasks has become a new paradigm. Low-Rank Adaptation (LoRA) significantly reduces the number of trainable parameters for fine-tuning. Although it has demonstrated commendable performance, updating parameters within a single scale may not be the optimal choice for complex downstream tasks.
in this paper, we extend the LoRA to multiple scales, dubbed as LoRA$^2$. 
We first combine orthogonal projection theory to train a set of LoRAs in two mutually orthogonal planes. Then, we improve the importance score algorithm, which reduce parameter sensitivity score calculations by approximately 98.5\%. By pruning singular values with lower importance scores, thereby enhancing adaptability to various downstream tasks. 
Extensive experiments are conducted on two widely used pre-trained models to validate the effectiveness of LoRA$^2$. 
Results show that it significantly reduces the number of trainable parameters to just 0.72\% compared to full fine-tuning, while still delivering highly impressive performance. Even when the parameters are further reduced to 0.17M, it still achieves comparable results to the baseline with 8 times more parameters. Our code is available here: \url{https://anonymous.4open.science/r/LoRA-2-5B4C}
\end{abstract}

\section{Introduction}

Large Language Models (LLMs) have become the cornerstone of NLP tasks \cite{devlin-etal-2019-bert,zhuang-etal-2021-robustly,he2021deberta,Radford2019LanguageMA}. The powerful emergent abilities \cite{wei2022emergent} enables LLMs to adapt to downstream tasks through fine-tuning. The simplest approach is to fine-tune all the parameters of LLMs \cite{Qiu_Sun_Xu_Shao_Dai_Huang_2020,zhuang-etal-2021-robustly}. 
\begin{figure}[!t]

\centerline{\includegraphics[width=1.0\columnwidth]{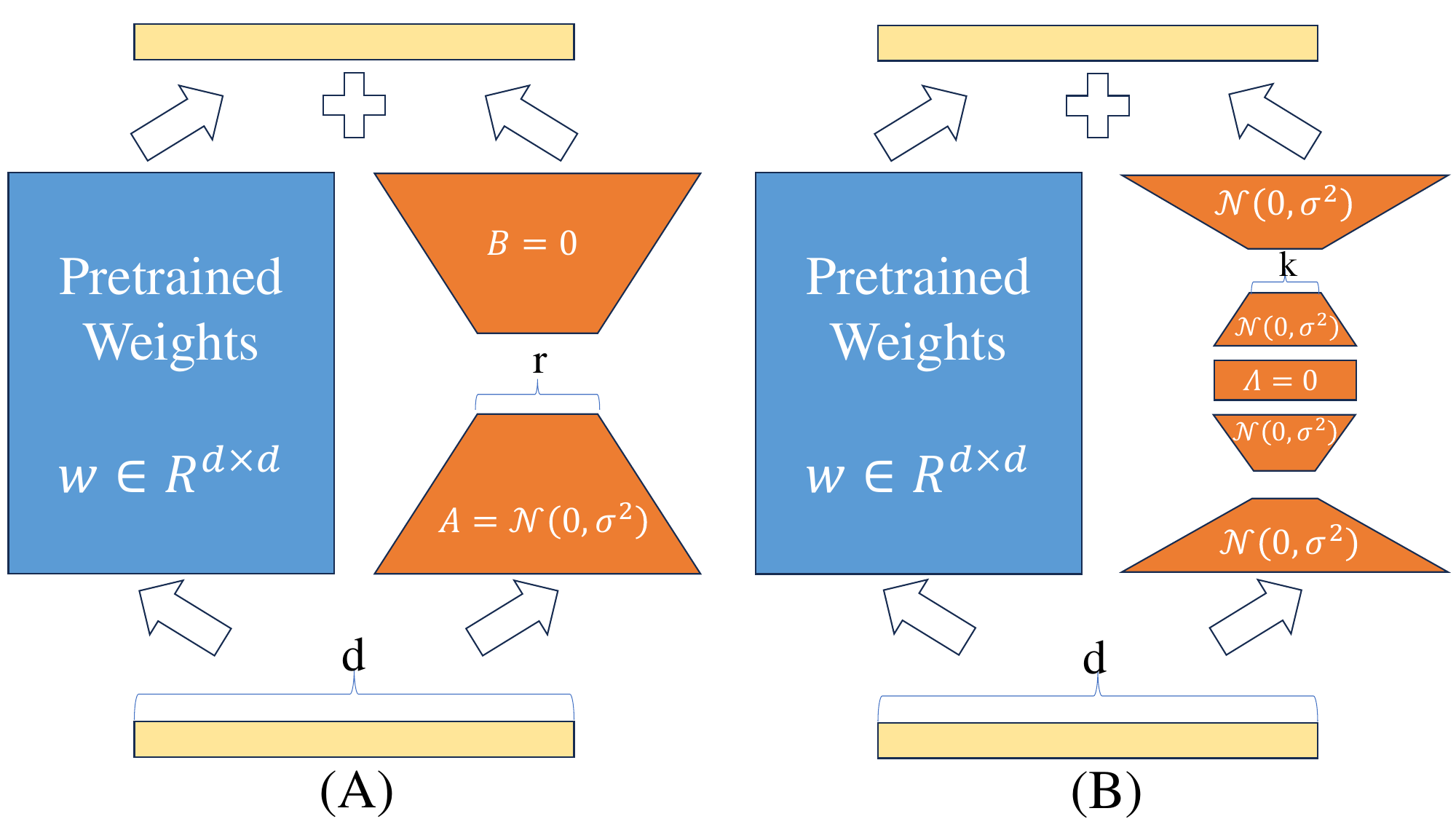}}
  \caption{Blue blocks represent frozen parameters, while orange represents trainable parameters. (A) LoRA only utilizes a set of low-rank matrices to approximate increments. (B) LoRA$^2$ trains a set of low-rank matrices in two mutually orthogonal planes.}
\end{figure}
\begin{figure*}[t]

\centerline{\includegraphics[width=\textwidth]{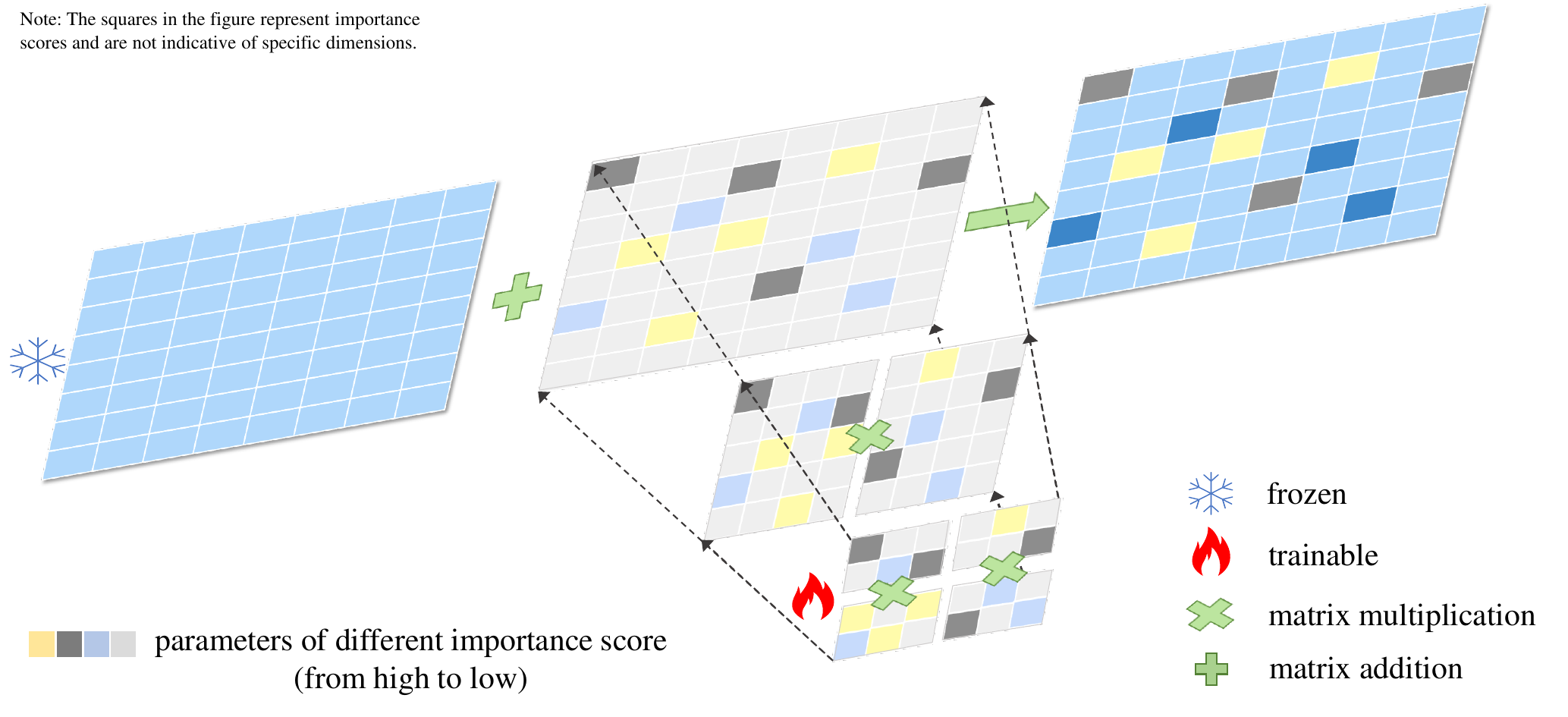}}
  \caption{An illustration of Multi-Scale Orthogonal Low-Rank Approximations (LoRA$^2$):  Based on the principle of orthogonal projection, We train a set of inherently orthogonal LoRAs on orthogonal planes as the incremental matrices.}
  \label{photo1}
\end{figure*}
However, as the model's parameters gradually expand, PaLM \cite{10.5555/3648699.3648939} contains up to 540 billion parameters; GPT-4 \cite{Achiam2023GPT4TR} contains up to 100 trillion parameters. The massive memory and time resources required to fine-tune all the parameters of these models are completely unacceptable.

To address this issue, LoRA \cite{hu2022LoRA} propose learning incremental updates to pre-trained weights through the product of two small matrices. LoRA avoids forward propagation latency caused by inserting additional neural modules while demonstrating stable performance. Compared to fine-tuning, only less than 0.5\% additional trainable parameters are needed, and training overhead can be reduced by up to 70\%. LoRA achieves performance comparable to or even better than fine-tuning. But LoRA still has limitations as it prespecifies the rank r of each incremental matrix identical.

In response to this issue, AdaLoRA \cite{zhang2023adaptive} dynamically allocates parameter budgets between weight matrices. Sensitivity scores are computed for all parameters of the matrices, which are aggregated into importance scores for singular values. The lowest-ranking singular values are pruned. Such operations enable adaptive manipulation of the rank. Although the performance of AdaLoRA is improved, but results in slower convergence.

In this paper, We first combine orthogonal projection theory to train a set of LoRAs on two different planes and make them orthogonal through dual regularization. The external regularizer minimizes the overlap of LoRA blocks, thereby enlarging the learning space of multi-scale LoRAs. The dual internal regularizers enhance the regularity of parameter updates, accelerating convergence. Similar to AdaLoRA, we adopt dynamic parameter budget allocation method. We extend the computation method of importance scores to adapt to the complex matrices of LoRA$^2$.
Due to the properties of matrix multiplication, where the rows of the preceding matrix must be multiplied by the columns of the succeeding matrix, we add the importance scores of the column matrix to the importance scores of the row matrix. But further discover through reasoning that since the calculation of each singular value's importance score includes the sensitivity scores of all parameters in the column matrix, the sensitivity scores of the column matrix have no impact on parameter pruning. Therefore, we exclude the column matrix calculations when computing importance scores.
LoRA$^2$ adapts to various downstream by pruning with importance scores.
Extensive experiments are conducted on various tasks and models to demonstrate the effectiveness of LoRA$^2$. Specifically, natural language understanding GLUE \cite{wang-etal-2018-GLUE} is evaluated using DeBERTaV3-base \cite{he2023debertav}. The findings, including qualitative and quantitative results, indicate that LoRA$^2$ outperforms existing methods. The contributions of this paper are as follows:
\begin{itemize}
\item[$\bullet$] We propose LoRA$^2$, a novel method that trains two internally LoRAs on orthogonal planes. It greatly increases the model's learnable space in a low-rank environment. Compared to full fine-tuning, it reduces the number of trainable parameters to 0.72\%.
\end{itemize}
\begin{itemize}
\item[$\bullet$] We improve the importance score algorithm to accommodate the structure of LoRA$^2$. It reduce the parameter sensitivity score calculations by approximately 98.5\% without causing any degradation in performance. By dynamically allocating parameter budgets based on importance scores, we achieve adaptability across various downstream tasks.
\end{itemize}
\begin{itemize}
\item[$\bullet$]Extensive experiments are conducted to demonstrate the effectiveness of our method. Particularly, our model could consistently outperform parameter-efficient baselines with fewer parameters on a wide range of downstream tasks.
\end{itemize}

\section{Related Work}
Parameter-Efficient Fine-Tuning (PEFT) is a strategy to adapt large-scale pre-trained models effectively and resource-efficiently for specific tasks. In fields such as NLP, pre-trained models like BERT and GPT-3 are highly favored due to their powerful representation learning capabilities. However, directly fine-tuning these models with all their parameters to fit new tasks often requires significant computational resources and time.

PEFT aims to overcome this challenge by updating only a small subset of crucial parameters in the model, rather than all millions or even billions of parameters. It achieves efficient and precise task transfer. Here are several major PEFT methods:

\textbf{Adapter Module} \cite{pmlr-v97-houlsby19a,li-liang-2021-prefix}: Inserting compact trainable modules (such as residual blocks) into various layers of the pre-trained model and fine-tuning only these adapters while keeping the original model parameters unchanged. This approach preserves pre-training knowledge while allowing adapters to capture task-specific information.

\textbf{Prompt Tuning} \cite{Sahoo2024ASS}: represents an efficient approach to adapt Transformer-based models to new tasks by optimizing only the prompt vector, rather than adjusting the model weights directly. This method involves inserting a fixed-length trainable vector, known as a prompt, at the input stage, which interacts with the standard fixed inputs to guide the model’s output generation in a task-specific manner. The key advantage of prompt tuning lies in its ability to adapt the model to new tasks with minimal disruption to the underlying model architecture. By merely adjusting the prompts, the approach avoids the computationally expensive and potentially destabilizing process of retraining large portions of the model. This results in a significant reduction in the number of parameters that need to be trained, making the process more efficient and scalable.

\textbf{LoRA} \cite{hu2022LoRA}: Fine-tuning by adding low-rank correction terms to the model weight matrices instead of directly updating the entire weight matrix. It utilizes low-rank approximation theory to effectively adjust the model's behavior with smaller parameter increments. formulated as:
\begin{equation}
  \label{eq:1}
W = W^{(0)} + \Delta  = W^{(0)} + BA,
\end{equation}

where $\Delta  \in  \mathbb{R}^{din\times dout}$, $A \in  \mathbb{R}^{r\times dout}$, and $B \in \mathbb{R}^{din\times r}$, with $r^{ \in (din, dout)}$. The dimensions of $din$ and $dout$ are the same as those of the pre-trained matrix $W$. During fine-tuning, only $A$ and $B$ are updated. The rank $r$ is chosen to be much smaller than the dimension of $W$. With less than 0.5\% additional trainable parameters, the training overhead can be reduced up to 70\%. Therefore, using LoRA to fit the incremental matrix increases the sparsity of the incremental matrix, and theoretically demonstrates the role of sparsity in model stability. As the compression ratio $p$ decreases, the upper bound also decreases. 
Ref. \cite{10.1609/aaai.v37i11.26505} derives Equation \ref{eq:5} based on the pointwise hypothesis stability (PHS) \cite{10.1145/567806.567809} to prove that using LoRA for fine-tuning implies better stability. 
\begin{align}
\label{eq:5}
 E_{S,i\sim U(n)}|\ell(\A(S^{i}),z_i)-\ell(\A(S),z_i)|\nonumber& \\
 \le \frac{2\rho^2 }{(\Lambda_{min}+2(1-p))n},
\end{align}\\
$A(S)$ is defined as the model parameters obtained by running algorithm $A$ on data $S$. $\Lambda_{\text{min}} = \min\{\Lambda_1, \dots, \Lambda_m\}$. $\ell(\cdot)$ represents the loss function. The variable $\rho$ represents this measure of stability, reflecting the maximum impact of input variations on the output in the loss function. Equation \ref{eq:5} demonstrates that as the sparsity parameter $p$ decreases, the upper bound also decreases. Therefore, sparse models are associated with better stability.

\textbf{AdaLoRA} \cite{zhang2023adaptive} dynamically allocates parameter budgets among weight matrices based on LoRA, and parametrizes the incremental updates to the pre-trained weight matrices using singular value decomposition:
\begin{equation}
  \label{eq:2}
W = W^{(0)} + \Delta = W^{(0)} + P \Lambda Q,
\end{equation}

where $P$ and $Q$ denote the left and right singular values, and the diagonal matrix $\Lambda$ contains the top $r$ largest singular values. Initially, $\Lambda$ is initialized to zero, and $P$ and $Q$ are randomly initialized with Gaussian distribution to ensure $P \Lambda Q = 0$ at the beginning of training. Specifically, the model retains only the top-scoring entries based on their newly proposed metric of sensitivity, heuristically constructed from the product of weight gradients.
Although the AdaLoRA method achieved incremental results in benchmark testing, it still exhibits significant redundancy in terms of parameters. The updates to the matrix are irregular, resulting in slow fitting speeds when encountering complex tasks. Here are two unresolved issues: (1) Under low-rank conditions, the model's performance is adversely affected, rendering it unable to adapt effectively to low-resource environments. (2) When $r=1$, the dynamic allocation of parameter budgets fails to work.

\section{Our Method}
\label{article3}
Our method consists of two main parts: (1) Multi-Scale Orthogonal Approximation. We use multi-scale orthogonal matrices to approximate the incremental matrix and minimize the overlap of LoRA through dual orthogonality constraints. (2) Complex Matrix Importance Pruning. We extend the orthogonality constraints to minimize spatial overlap. 


\subsection{Multi-Scale Orthogonal Approximation}
The overall pipeline of LoRA$^2$ is illustrated in Figure \ref{photo1}. We utilize Singular Value Decomposition (SVD) to project the parameter increment matrix $\Delta$ onto a mutually orthogonal plane. Aiming to reduce model complexity and enhance computational efficiency, while preserving crucial information as much as possible through the singular value matrix $\Lambda$. Then, we iteratively apply low-rank matrices. As shown in Figure \ref{photo1}, training two low-rank matrices within two orthogonal plane.  Ultimately, the parameter increment matrix is added to the pretrained parameters to adapt to various downstream tasks, offering a plug-and-play capability. Our forward propagation proceeds as follows:
\begin{align}
  \label{eq:4}
W = W^{(0)} + \Delta = W^{(0)} + P \Lambda Q \nonumber& \\
= W^{(0)} + (uv) \Lambda (\mathcal{UV}),
\end{align}

where $v \in\mathbb{R}^{(din,k)}$, $u \in\mathbb{R}^{(k,r)}$, $V \in\mathbb{R}^{(k,dout)}$ and $U \in\mathbb{R}^{(r,k)}$ are two sets of orthogonal LoRA matrices. 
The variable $k$ is a hyperparameter used to determine the dimension to which the data is projected. The matrix $\Lambda$ is a diagonal matrix. $\Lambda$ is initialized with zero while $u$, $v$, $\mathcal{U}$ and $\mathcal{V}$ adopt a random Gaussian initialization to ensure $\Delta = 0$ at the beginning of training. LoRA$^2$ employs regularization terms to increase the orthogonality of the matrices. This is similar to the AdaLoRA, which converts a constrained optimization problem into an unconstrained version, thereby enlarging the representational space of the matrices. To utilize regularization terms in LoRA$^2$, we expand the scope of the regularization terms. We further introduce regularizers between the $uv$ and $\mathcal{UV}$ matrices:
\begin{equation}
  \label{eq:6}
    \begin{cases}
        R(\mathcal{P, Q}) = \left\| P^T P - I \right\|_{F}^{2} + \left\| Q Q^T - I \right\|_{F}^{2}\\
        R(\mathcal{U, V}) = \left\| \mathcal{U}^T \mathcal{U} - I \right\|_{F}^{2} + \left\| \mathcal{V V}^T - I \right\|_{F}^{2},\\
        R(u, v) = \left\| u^T u - I \right\|_{F}^{2} + \left\| v v^T - I \right\|_{F}^{2}
    \end{cases}
\end{equation}

dual regularization helps enhance the optimization stability of the matrix. By minimizing the overlap in LoRA, the learning space of the method is expanded. In Section \ref{Orthogonal Constraint Analysis}, we present an ablation study to demonstrate the effectiveness of our regularization approach.
\begin{algorithm}[ht!]
 	\caption{{Pruning Algorithm of LoRA$^2$}} 
 	\label{alg:our_algorithm}
 	\begin{algorithmic}[1]
 		\STATE {{\bfseries Input:} Dataset $ \mathcal{D} $; total iterations $ T $. }  
 		\FOR{$ t = 1, \dots, T $}  
 		\STATE Sample a mini-batch from $\mathcal{D}$;
            \STATE Compute gradient $ \nabla\LL(v, \Lambda, \mathcal{V}) $;
 		\STATE Compute $\overline{I}^{(t)}_{ij}$ and $\overline{U}^{(t)}_{ij}$as (\ref{eq:9} and \ref{eq:example});
 		\STATE Compute Importance $\mathbb{I}(\Lambda,i)$as (\ref{eq:13});
 		\STATE Compute threshold $\mathcal{I}$, for $ i=1,\dots,r $;
            \STATE Mask $(\mathbb{I}(\Lambda,i) < \mathcal{I}$).  
 		\ENDFOR 
            \STATE \textbf{Output:}  { The fine-tuned parameters $(\Lambda, i)$.} 
	\end{algorithmic}
\end{algorithm}
\subsection{Complex Matrix Importance Pruning}
In LoRA$^2$, the matrix $\Lambda$ is iteratively pruned to adjust the rank after each gradient descent step.  We adopt the same rank pruning method as AdaLoRA, by applying the SVD-based adaptation to every weight matrix including W$_q$, W$_k$, W$_v$, W$_{f1}$ and W$_{f2}$ of each transformer layer. We summarize the detailed algorithm in Algorithm~\mbox{\ref{alg:our_algorithm}}.

where $0 <\beta_1, \beta_2 < 1$. $\overline{I}^{(t)}$ is the sensitivity smoothed by an exponential moving average, and $\overline{U}^{(t)}$ is the uncertainty term quantified by the local variation between $\overline{I}^{(t)}$ and $I^{(t)}$. They then define the importance as the product of $\overline{I}^{(t)}$ and $\overline{U}^{(t)}$.

since LoRA$^2$ trains two LoRA blocks at multiple scales, there are four matrices related to singular values. The dimensions of matrices $u$ and $\mathcal{U}$ are the same as those of the singular value matrix $\Lambda$, whereas the dimensions of $v$ and $\mathcal{V}$ differ from $\Lambda$. Since the number of columns in matrices $u$ and $\mathcal{U}$ and the number of rows in matrices $v$ and $\mathcal{V}$ are all equal to the hyperparameter $k$, and each parameter in matrices $u$ and $\mathcal{U}$ needs to be multiplied by matrices $v$ and $\mathcal{V}$. We average  the importance scores of each column in $u$ and $\mathcal{U}$ and add it to the importance scores of each row in $v$ and $\mathcal{V}$. Our total importance score calculation formula is as follows:
\begin{equation}
  \label{eq:11}
C(K, v, ij) = \sum_{j=1}^{K}S_{ij}(v),
\end{equation}
\begin{equation}
  \label{eq:12}
\overline{S}(K, u) =\sum_{i=1}^{K}\sum_{j=1}^{din}S_{ji}(u),
\end{equation}


\begin{align}
  \label{eq:4}
I(\Lambda,i) = S(\Lambda, i) + \frac{C(K, v, ij)+\overline{S}(K, u)}{K} \nonumber&\\
+ \frac{C(K, \mathcal{V}, ij)+\overline{S}(K,\mathcal{U})}{K},
\end{align}
Equation \ref{eq:11} represents the sum of row sensitivity scores for a low-rank matrix $v$ with $k$ rows. Equation \ref{eq:12} represents the sum of the column sensitivity scores for a low-rank matrix $u$ with $k$ columns. $I(\Lambda,i)$ denotes the importance score for the $i^{th}$ singular value $\Lambda$, calculated by averaging the sums of row sensitivity scores from matrices $v$ and $\mathcal{V}$, and column sensitivity scores from matrices $u$ and $\mathcal{U}$. Then adding the inherent sensitivity score of the singular value $\Lambda$. Since the importance score of each singular value $\Lambda$ includes the average of the sensitivity scores of all parameters in matrices $u$ and $\mathcal{U}$, the ranking is unaffected. 
Thus, in practical terms, we can disregard the sensitivities of matrices $u$ and $\mathcal{U}$. Surprisingly, this approach reduces the calculation of parameter sensitivity scores by approximately 98.5\%. Our method for calculating importance scores is as follows:
\begin{align}
  \label{eq:13}
\mathbb{I}(\Lambda,i) = S(\Lambda, i) + \frac{C(K, v, ij)}{K}+\frac{C(K, \mathcal{V}, ij)}{K}.
\end{align}

\begin{table*}[!ht]
    \centering
    \scalebox{0.74}{
    \renewcommand\arraystretch{1.35}
    \begin{tabular}{l|l|c|ccccccccc}
    \toprule
        \textbf{Model}&\textbf{Method} & \textbf{\#Params} & \textbf{CoLA} & \textbf{SST-2} & \textbf{MRPC} & \textbf{QQP} & \textbf{STS-B} & \textbf{MNLI} & \textbf{QNLI} & \textbf{RTE} & \textbf{Avg.}  \\  
        \midrule
        DeB$^{V3}_{base}$&Fine-Tune  &184M  & 69.21 & 95.64 &89.22 & 92.05/89.31 & 91.59 &89.98/89.95 &93.78 &82.49 &87.82 \\ 
        \midrule
        DeB$^{V3}_{base}$&Bitfit &0.1M &68.70 &94.38 & 87.16 & 87.86/84.20 &89.71 &87.45/87.45 &91.90 & 76.12 &85.18 \\
        DeB$^{V3}_{base}$&AdaLoRA $(r=2)$ &0.32M & 70.04 &\textbf{95.80} & \underline{90.44} & \textbf{91.78}/\textbf{89.16}  &\underline{91.63} &\textbf{90.66}/\textbf{90.70}  &\textbf{94.49}&\underline{87.36} &\underline{88.86} \\
        DeB$^{V3}_{base}$&LoRA $(r=1)$ &0.17M &68.60 &94.95 &88.24 &91.20/88.37 &91.41 &90.09/90.28  &93.35 &81.29&87.23 \\
        DeB$^{V3}_{base}$&SoRA $(r=1)$ &0.12M &\underline{70.24} &95.14 &89.22 &91.52/\underline{88.73} &91.41 &90.08/\underline{90.41} &93.43 &83.02&87.85 \\
        DeB$^{V3}_{base}$&LoRA$^2$$(k=1)$ &0.17M &\textbf{70.63} &\underline{95.64} &\textbf{90.93} &\underline{91.62}/88.05 &\textbf{91.69} &90.19/90.38  &\underline{93.92} &\textbf{89.53} &\textbf{89.06} \\
        \midrule
        DeB$^{V3}_{base}$&AdaLoRA $(r=8)$ &1.27M & 71.45 &\textbf{96.1} & 90.69 & 92.23/\underline{89.74}  &91.84 &\textbf{90.76}/\textbf{90.79}  &\textbf{94.55}&\underline{88.09} &\underline{89.31} \\
        DeB$^{V3}_{base}$&LoRA $(r=8)$ &1.33M &69.73 &95.57 &89.71 &91.95/89.26 &91.86 &\underline{90.47}/90.46  &93.76 &85.32&88.38 \\
        DeB$^{V3}_{base}$&SoRA $(r=8)$ &0.91M &\underline{71.48} &95.64 &\textbf{91.98} &\textbf{92.39}/\textbf{89.87} &\textbf{92.22} &90.35/\underline{90.38} &\underline{94.28} &87.77&89.36 \\
        DeB$^{V3}_{base}$&LoRA$^2$$(k=8)$ &1.33M &\textbf{72.30} &\underline{95.76} &\underline{91.42} &\underline{92.25}/\underline{89.74} &\underline{91.92} &90.43/90.23 & \underline{94.28} &\textbf{88.45} &\textbf{89.68} \\
    \bottomrule
    \end{tabular}}
    \caption{Test results of LoRA$^2$ and other baselines on the GLUE benchmark. We report the matched and mismatched accuracy for MNLI, Matthew's correlation for CoLA, Pearson correlation for STS-B, and accuracy for other tasks. Higher is better for all metrics. We use the same hyperparameters, which are specified in Appendix \ref{A}. The optimal training values are used as results. We denote the best result in \textbf{bold} and \underline{underline} the second-best result.}
    \vspace{-0.4cm}
    \label{tab:main}
\end{table*}



\section{Experiments}

We implement LoRA$^2$ for fine-tuning DeBERTaV3-base \cite{he2023debertav} and RoBERTa-large \cite{zhuang-etal-2021-robustly}, We evaluate the effectiveness of the proposed algorithm on natural language understanding tasks from the GLUE benchmark \cite{wang-etal-2018-GLUE}.
\subsection{Experimental Settings}
\textbf{Implementation Details.}  We use PyTorch \cite{10.5555/3454287.3455008} to implement all the algorithms. Our implementation is based on the publicly available Huggingface Transformers3 \cite{wolf-etal-2020-transformers} code-base. All the experiments about DeBERTaV3-base \cite{he2021deberta} are conducted on NVIDIA 4090 laptop GPU and experiments about RoBERTa-large \cite{zhuang-etal-2021-robustly} are conducted on NVIDIA A800 GPU. Since the total number of parameters in our model is primarily controlled by the hyperparameter $K$, the rank $R$ has almost no impact on the total amount of trainable parameters. Therefore, we keep the $K$ of LoRA$^2$ consistent with the R of the baselines. This allows for comparisons between models with a similar amount of trainable parameters. We mainly use DeBERTaV3-base \cite{he2023debertav} as the backbone model. Additionally, we also use RoBERTa-large \cite{zhuang-etal-2021-robustly} for analysis.\\

\noindent\textbf{Baselines.}  Our baselines comprise full-parameter fine-tuning and other well-recognized parameterefficient methods, including Bitfit \cite{zhang2022platon} ,LoRA \cite{hu2022LoRA} , AdaLoRA \cite{zhang2023adaptive} and SoRA \cite{ding-etal-2023-sparse}.\\


\noindent\textbf{Datasets.}  For evaluation, we adopt the GLUE benchmark \cite{wang-etal-2018-GLUE}, a widely recognized benchmark for natural language understanding, including CoLA \cite{warstadt-etal-2019-neural}, SST-2 \cite{socher-etal-2013-recursive}, MRPC \cite{dolan-brockett-2005-automatically}, QQP \cite{wang-etal-2018-GLUE}, STS-B \cite{wang-etal-2018-GLUE}, MNLI \cite{williams-etal-2018-broad}, QNLI \cite{rajpurkar-etal-2016-squad} and RTE \cite{10.1007/11736790_9,giampiccolo-etal-2007-third,Bentivogli2011TheSP}. 

\subsection{Results}
We first conduct an evaluation on GLUE benchmark. The experimental performance of LoRA$^2$ and other baselines are shown in Table \ref{tab:main}. 
Our method results indicate that LoRA$^2$ consistently outperforms the baselines in most tasks. For instance, on the RTE, the accuracy of LoRA$^2$ reaches 89.53\%, which is 2.17\% higher than AdaLoRA $(r=2)$. On average, under the condition of $K/R=1$, LoRA$^2$ outperforms LoRA and SoRA on the GLUE benchmark by 2.03\% and 1.41\%. When the parameter amount increases to $K/R=8$, the performance of LoRA$^2$ further improves, exceeding LoRA and SoRA by 1.29\% and 0.31\% on average.
Specifically, even when comparing LoRA$^2$ ($k=1$) to other baselines with ($r=8$), it still slightly outperforms the baselines.
We also conduct an experiment on RoBERTa-large \cite{zhuang-etal-2021-robustly} to compare the performance of larger models. LoRA$^2$ exhibits remarkable capabilities. It achieves results comparable to the 335M parameters (full fine-tuning) while using only 0.4M parameters, thereby achieving a compression rate of 99.97\%. 
Further comparison between LoRA$^2$ and LoRA reveals that our method improves performance by 2\% while reducing parameters by 0.37M. These outcomes confirm that LoRA$^2$ maintains robust performance in large-scale pre-trained models, effectively demonstrating the versatility of LoRA$^2$. 
The results prove that the advantage of LoRA$^2$ is consistent across different model sizes, achieving results comparable to a baseline with four times the parameter quantity.

\begin{figure*}[!t]

\centerline{\includegraphics[width=\textwidth]{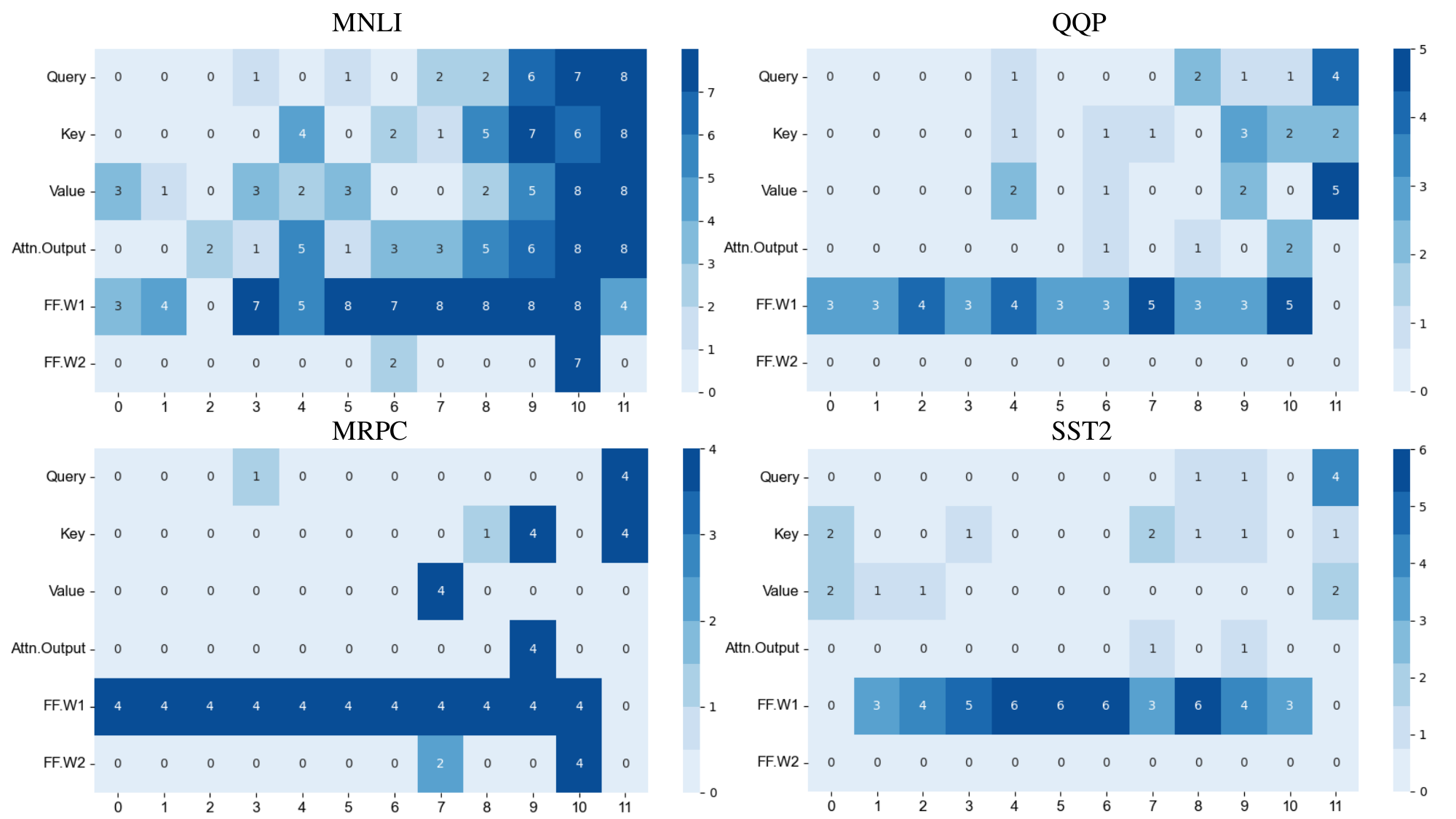}}
  \caption{The final rankings after training with LoRA$^2$ $(r=8)$ on four datasets (i.e., MNLI, QQP, MRPC, and SST2). The X-axis is the index of DeBERTaV3-base layers, and the Y-axis indicates the different layers to which LoRA$^2$ is applied. The lighter the color, the lower the degree of pruning.}
\label{photo2}
\end{figure*}

\subsection{Rank Analysis}
For a fixed model, it is inevitable that adapting to different downstream tasks can be challenging. At the same time, the importance of the model's parameters varies. Some parameters contribute minimally to the final outcomes, not only failing to enhance the model's capabilities but also impacting the convergence speed. Therefore, adaptively adjusting the parameter budget is crucial. In this section, we will show and analyze the final rankings of parameters after stable training on four datasets using the LoRA$^2$ method, as illustrated in Figure~\mbox{\ref{photo2}}. The analysis results reveal that even after fitting with a low-rank matrix, the predefined rank still goes far beyond what is required for fine-tuning specific tasks. More than half of the ranks have minimal impact on the final outcome. Comparing different tasks, it is evident that MNLI requires the highest rank, while SST2 demands the lowest. Further analysis reveals that extensive pruning is needed for the top layers, with only the parameters in the FF.W1 layer being significant. In contrast, the demand for incremental parameters increases for the lower layers, demonstrating a clear heavy-tail structure. This phenomenon also indicates that using a constant parameter budget negatively affects the fine-tuning results, necessitating a case-by-case consideration.

\begin{table}[!htbp]
    \centering
    \scalebox{0.88}{
    \tabcolsep=0.35cm
    \begin{tabular}{l|ccc}
    \toprule
        \textbf{Datasets}& \textbf{AdaLoRA} & \textbf{LoRA$^2$}\\
        \midrule
        \textbf{CoLA}& 1.01s/it & 1.14\\
        \textbf{SST-2}& 1.04 & 1.19\\
        \textbf{MRPC}& 1.45 &1.68\\
        \textbf{QQP}& 1.44 & 1.64\\
        \textbf{STS-B} & 1.05 & 1.23\\
        \textbf{MNLI}& 1.31 & 1.50\\
        \textbf{RTE}& 1.44 & 1.65\\
        \midrule
        \textbf{Avg.}&335M & 0.8M\\
    \bottomrule
    \end{tabular}}
    \caption{The average training time per epoch on six datasets. For each task, the experiments with AdaLoRA and LoRA$^2$ have the same batch size 32.}
    \vspace{-0.2cm}
    \label{time}
\end{table}
\begin{table*}[htbp]
    \centering
    \scalebox{0.88}{
    \tabcolsep=0.35cm
    \begin{tabular}{l|ccccc}
    \toprule
        & \textbf{MRPC}& \textbf{STS-B}& \textbf{MNLI}& \textbf{RTE}& \textbf{Avg.}\\
        \midrule
        \textbf{LoRA$^2_{(uv)}$}&90.20&91.61&90.27 &89.17 &90.31\\
        \textbf{LoRA$^2_{(\mathcal{UV})}$}&90.20&91.57&90.23 &89.17 &90.29\\
        \textbf{LoRA$^2_{(uv\&\mathcal{UV})}$}&90.44&91.61&90.33  &89.53 &90.48\\
        \textbf{LoRA$^2_{(P\&Q)}$}&90.44&91.59&90.33 &88.81 &90.29\\
        \textbf{LoRA$^2_{(All)}$}&\textbf{90.93}& \textbf{91.69} &\textbf{90.38}&\textbf{89.53} &\textbf{90.63}\\
    \bottomrule
    \end{tabular}}
    \caption{Results using different regularization methods. LoRA$^2_{(uv)}$ indicates the application of orthogonal constraints to the matrix $uv$ of LoRA$^2$. $All$ represents the simultaneous application of orthogonal constraints to $uv$, $\mathcal{UV}$, and $PQ$.}
    \vspace{-0.2cm}
    \label{orthogonal}
\end{table*}
\begin{table*}[!htbp]
    \centering
    \scalebox{0.84}{
    \tabcolsep=0.35cm
    \begin{tabular}{l|c|ccccc}
    \toprule
        \textbf{Method}&\textbf{\#Params}&\textbf{CoLA}& \textbf{MRPC}& \textbf{STS-B}& \textbf{RTE}& \textbf{Avg.}\\
        \midrule
        \textbf{LoRA$^2$$_{(Q,K)}$}&37.4k &68.66&89.46&\textbf{91.69}&88.09&84.53\\
        \textbf{LoRA$^2$$_{(Q,V)}$}&37.4k &67.70&89.95&91.66&84.84&83.54\\
        \textbf{LoRA$^2$$_{(Q,K,V)}$}&56.2k &70.44&89.46&91.04&86.28&84.31\\
        \textbf{LoRA$^2$$_{(ALL)}$}&167.6k &\textbf{70.63}& \textbf{90.93}& \textbf{91.69}&\textbf{89.53}&\textbf{85.70}\\
    \bottomrule
    \end{tabular}}
    \caption{The results of applying LoRA$^2$ $(k=1)$ to different layers. $ALL$ represents the output of the query, key, value attention and FF layers respectively. \textbf{\#Params} refers to the number of trainable parameters.}
    \vspace{-0.2cm}
    \label{test}
\end{table*}

\subsection{Orthogonal Constraint Analysis}
\label{Orthogonal Constraint Analysis}
We evaluate five orthogonal methods on the GLUE benchmark. Experiments use the same hyperparameters, with only the method of orthogonal constraints changing. The experimental results are shown in Table \ref{test}. Applying orthogonal constraints simultaneously to $uv\&\mathcal{UV}\&PQ$ achieves the best results, outperforming the other three orthogonal methods by about 0.3\%. Further analysis shows that applying orthogonal constraints to either $uv\&\mathcal{UV}$ or $PQ$ yields similar results, and applying orthogonal constraints only to $\mathcal{UV}$ yields the worst results. We believe that constraining only $\mathcal{UV}$ without constraining $PQ$ in LoRA$^2$ leads to spatial overlap in the training of Multi-Scale LoRA, resulting in reduced learning space. In contrast, applying dual constraints maximizes the orthogonality between matrix planes, thereby maximizing the learnable space.

\subsection{Applying LoRA$^2$ to Different Weights}
In this section, we apply LoRA$^2$ $(k=1)$ to different types of weight matrices to determine how to achieve optimal performance on downstream tasks. We employ the DeBERTaV3-base model for fine-tuning tasks on the CoLA, MRPC, STS-B, and RTE datasets. As shown in Table \ref{orthogonal}, on the STS-B dataset alone, the results of applying LoRA$^2$ only to the $Q$ and $K$ matrices are equally excellent. However, for other tasks, applying LoRA$^2$ to all weight matrices yields the best results, with an average performance lead of over 1\%. This is largely consistent with the situation for LoRA. The results suggest that applying LoRA$^2$ to all weight matrices can be a beneficial strategy.
\section{Conclusion}
We propose a multi-scale low-rank approximation named LoRA$^2$, an innovative approach for efficiently fine-tuning large pretrained language models. Building on the basis of SVD, we train LoRAs at multiple scales on mutually orthogonal planes. By dynamically allocating parameter budgets through pruning, LoRA$^2$ adapts to various downstream tasks. We change the importance score algorithm to accommodate the structure of LoRA$^2$. It reduce the parameter sensitivity score calculations by approximately 98.5\% without causing any degradation in performance. We conduct extensive experiments in the NLP domain, and LoRA$^2$ achieves performance close to baselines with eight times the number of parameters, demonstrating that LoRA$^2$ surpasses existing methods.
\section*{Limitations}
Although LoRA$^2$ has demonstrated surprising performance on NLP datasets, our research still has some acknowledged limitations. However, recent studies have shown that methods for parameter-efficient fine-tuning can also be applied in the cross-modal domain, and the performance of LoRA$^2$ in the multimodal field is currently unknown. Additionally, our method only evaluates the fine-tuning results of dual LoRA at multiple scales. For more multi-scale LoRA, we intend to conduct experiments to verify its performance.


\bibliography{custom.bib}
\clearpage
\newpage
\appendix

\section{Datasets}
\label{A}

For evaluation, we adaopt the GLUE benchmark \cite{wang-etal-2018-GLUE}, including CoLA \cite{warstadt-etal-2019-neural}, SST-2 \cite{socher-etal-2013-recursive}, MRPC \cite{dolan-brockett-2005-automatically}, QQP \cite{wang-etal-2018-GLUE}, STS-B \cite{wang-etal-2018-GLUE}, MNLI \cite{williams-etal-2018-broad}, QNLI \cite{rajpurkar-etal-2016-squad} and RTE \cite{10.1007/11736790_9,giampiccolo-etal-2007-third,Bentivogli2011TheSP}. We present the dataset statistics of GLUE in the following table \ref{tab:data detail}.

\begin{table}[!h]
    \centering
    \scalebox{0.77}{ 
    \renewcommand\arraystretch{1.5}
    \begin{tabular}{l|ccccc} 
    \toprule
      \textbf{Dataset} &  \textbf{Metric}  &  \textbf{\#Train} & \textbf{\#Valid} & \textbf{\#Test}& \textbf{\#Label}  \\ \midrule
        CoLA & Mcc&8.5k &1,043 &1,063 &2  \\
        SST-2& Acc &67k &872 &1.8k &2 \\
        MRPC & Acc&3.7k &408 &1.7k &2\\
        QQP  & Acc/F1&364k &40.4k &391k &2 \\
        STS-B& Corr&5.7k &1.5k &1.4k &1 \\
        MNLI & Acc(m/mm)&393k &20k &20k&3 \\
        QNLI & Acc&105k &5.5k &5.5k &2 \\
        RTE  & Acc&2.5k &277 &3k &2\\
    \bottomrule
    \end{tabular}}
    \caption{ Dataset Sizes and Evaluation Metrics in the GLUE Benchmark. "Mcc," "Acc," "F1," and "Corr" denote the Matthews correlation coefficient, accuracy, F1 score, and Pearson correlation coefficient, respectively. "Acc(m/mm)" indicates accuracy results for matched and mismatched datasets within MNLI.}
    \label{tab:data detail}
\end{table}
\section{Sparse Regularization Theory}

By using progressive projection matrices, we further increase the compression ratio $p$ of the parameters. Additionally, this enhances the stability of the fine-tuning process. Equations \ref{eq:15}\cite{10.1609/aaai.v37i11.26505} theoretically demonstrate the role of sparsity in model stability. As the compression ratio $p$ decreases, the upper bound also decreases. Therefore, a sparser model implies better stability.
\begin{align}
\label{eq:15}
 \E_{S,i\sim U(n)}[|\ell(\A(S^{i}),z_i)-\ell(\A(S),z_i)|]\nonumber& \\
 \le \frac{2\rho^2 }{(\Lambda_{min}+2(1-p))n},
\end{align}\\
$\E_{S,i\sim U(n)}[\cdot]$ is Pointwise Hypothesis Stability (PHS)\cite{10.1145/567806.567809}  which focuses on analyzing the change of model output after a training sample is removed. $\Lambda_{\text{min}} = \min\{\Lambda_1, \dots, \Lambda_m\}$. $\ell(\cdot)$ represents the loss function. The variable $\rho$ represents this measure of stability, reflecting the maximum impact of input variations on the output in the loss function. 

\section{Orthogonal Projection Theory}
It is a fundamental concept in linear algebra with applications across various fields including machine learning, statistics, and computer graphics \cite{lay2016linear}. This theory revolves around the idea of projecting a vector onto a subspace in a way that minimizes the distance between the vector and the subspace, effectively finding the closest approximation within that subspace.

Mathematically, consider a vector $u$ in $R_n$ and a subspace $\mathbb{V}$ spanned by vectors $ \{v_1, v_2, \ldots, v_k\} $. The orthogonal projection of u onto $\mathbb{V}$, denoted as $\mathbb{P}_\mathbb{V}(\mathbf{u})$, is given by:
\begin{equation}
\label{eq:23}
\mathbb{P}_\mathbb{V}(\mathbf{u}) = \sum_{i=1}^k \frac{\mathbf{u} \cdot \mathbf{v}_i}{\mathbf{v}_i \cdot \mathbf{v}_i} \mathbf{v}_i
\end{equation}
LoRA$^2$ enhances the model's learning and representational capabilities by training two mutually orthogonal LoRA blocks. This design allows each LoRA block to capture information in different dimensions, thereby reducing information overlap and increasing the overall efficiency and effectiveness of the model. Additionally, the orthogonal training strategy helps prevent overfitting, making the model more robust when faced with new data.

\end{document}